\documentclass[11pt,a4paper]{article}
\usepackage{times}
\usepackage{tgtermes}
\usepackage[T1]{fontenc}
\usepackage[utf8]{inputenc}
\usepackage[french]{babel}
\usepackage{fancyhdr}
\usepackage{setspace}
\usepackage{graphicx}
\usepackage{color}
\usepackage{xcolor}
\usepackage{soul}
\usepackage{caption}
\usepackage{subcaption}
\usepackage{eurosym}
\usepackage{url}
\usepackage{siunitx}
\usepackage{titlesec}
\titleformat{\section}
  {\normalfont\fontsize{16}{20}\bfseries}{\thesection}{1em}{}
\titleformat{\subsection}
  {\normalfont\fontsize{16}{20}\bfseries}{\thesubsection}{1em}{}
\titlespacing*{\section}
{0pt}{2.ex plus 1ex minus .2ex}{.0ex}
\titlespacing*{\subsection}
{0pt}{0.ex}{.0ex}

\begin{document}

\begin{center}
\begin{spacing}{2.05}
{\fontsize{20}{20}
\bf
Cinématique d'une Prothèse de Main Myoélectrique \\
Accessible avec Actionneur Unique
et Rétropulsion Passive du Pouce
}
\end{spacing}
\end{center}
\vspace{-1.25cm}
\begin{center}
{\fontsize{14}{20}
\bf
C. BUTIN\textsuperscript{a}, D. CHABLAT\textsuperscript{b}, Y. AOUSTIN\textsuperscript{c}, D. GOUAILLIER\textsuperscript{d}\\
\bigskip
}
{\fontsize{12}{20}
a. ORTHOPUS, Nantes \& Nantes Université, École Centrale de Nantes, CNRS, LS2N, UMR 6004, F-44000, Nantes, France, come.butin@orthopus.com \\
b. Nantes Université, École Centrale de Nantes, CNRS, LS2N, UMR 6004, F-44000, Nantes, France, damien.chablat@cnrs.fr \\
c. Nantes Université, École Centrale de Nantes, CNRS, LS2N, UMR 6004, F-44000, Nantes, France, yannick.aoustin@univ-nantes.fr \\
d. ORTHOPUS, Nantes, david.gouaillier@orthopus.com \\
}
\end{center}

\vspace{10pt}

{\fontsize{16}{20}\bf R\'esum\'e~:}
\medskip
\textit{
Ce travail propose une nouvelle cinématique de prothèse de main myoélectrique disposant d'un actionneur unique, permettant de réaliser la prise tridigitale mais aussi la prise latérale. En s'inspirant des prothèses tridigitales, qui sont plus simples, plus robustes et moins chères que les prothèses polydigitales, cette nouvelle cinématique a pour but de proposer une prothèse accessible (abordable, facile à utiliser, robuste, réparable). Des câbles sont utilisés à la place d'une bielle rigide pour transmettre le mouvement entre les doigts supérieurs et le pouce. Les méthodes et choix de conception sont détaillés dans cet article. Pour conclure, l'évaluation d'un prototype par un utilisateur expérimenté permet une première critique des résultats.
}

\vspace{20pt}

{\fontsize{16}{20}\bf Abstract:}
\bigskip
\textit{
This work proposes a new kinematics of a myoelectric hand prosthesis with a single actuator, allowing to realize the tridigital grip but also the lateral grip. Inspired by tridigital prostheses, which are simpler, more robust and less expensive than polydigital prostheses, this new kinematics aims at proposing an accessible prosthesis (affordable, easy-to-use, robust, easy-to-repair). Cables are used instead of a rigid rod to transmit the movement bewteen the upper fingers and the thumb. The methods and design choices are detailed in this article. To conclude, the evaluation of the prototype by an experimented user leads to a first discussion of the results.
}

\vspace{28pt}

{{\fontsize{14}{20}
\bf
Mots clefs~:} Prothèse de Main; Myoélectrique; Accessible; Cinématique.
}

{{\fontsize{14}{20}
\bf
Keywords:} Hand Prosthesis, Myoelectric, Accessible, Kinematics.
}

\vspace{50pt}

\section{Introduction}
\medskip

La conception d’une prothèse de main motorisée est un problème complexe, et qui reste un défi technique bien que de plus en plus étudié. D'une part, les prothèses commerciales à prise tridigitale (1 degré d'actionnement) sont connues depuis plusieurs décennies et sont reconnues pour leur simplicité et leur robustesse. Cependant, leur prix élevé (environ ~$8000$ \euro \footnote{Prix remboursés par la Sécurité Sociale en France~: \url{http://www.codage.ext.cnamts.fr/codif/tips//chapitre/index_chap.php?p_ref_menu_code=625&p_site=AMELI}}) ne leur permet pas une accessibilité suffisante et leur prise unique peut être vue comme limitative pour l’utilisateur. De nouvelles prothèses sont développées et arrivent sur le marché, avec une volonté commune~: permettre des mouvements plus complexes et réalistes. Cette volonté, bien que très louable, freine néanmoins fortement l'accessibilité de ces mains robotiques, de part leur prix bien plus élevé (environ ~$30000$ \euro \footnotemark[1]), leur fragilité et parfois leur lenteur.\par

\begin{figure}[!ht]
    \centering
    \begin{subfigure}[b]{0.35\textwidth}
        \centering
        \includegraphics[width=\textwidth]{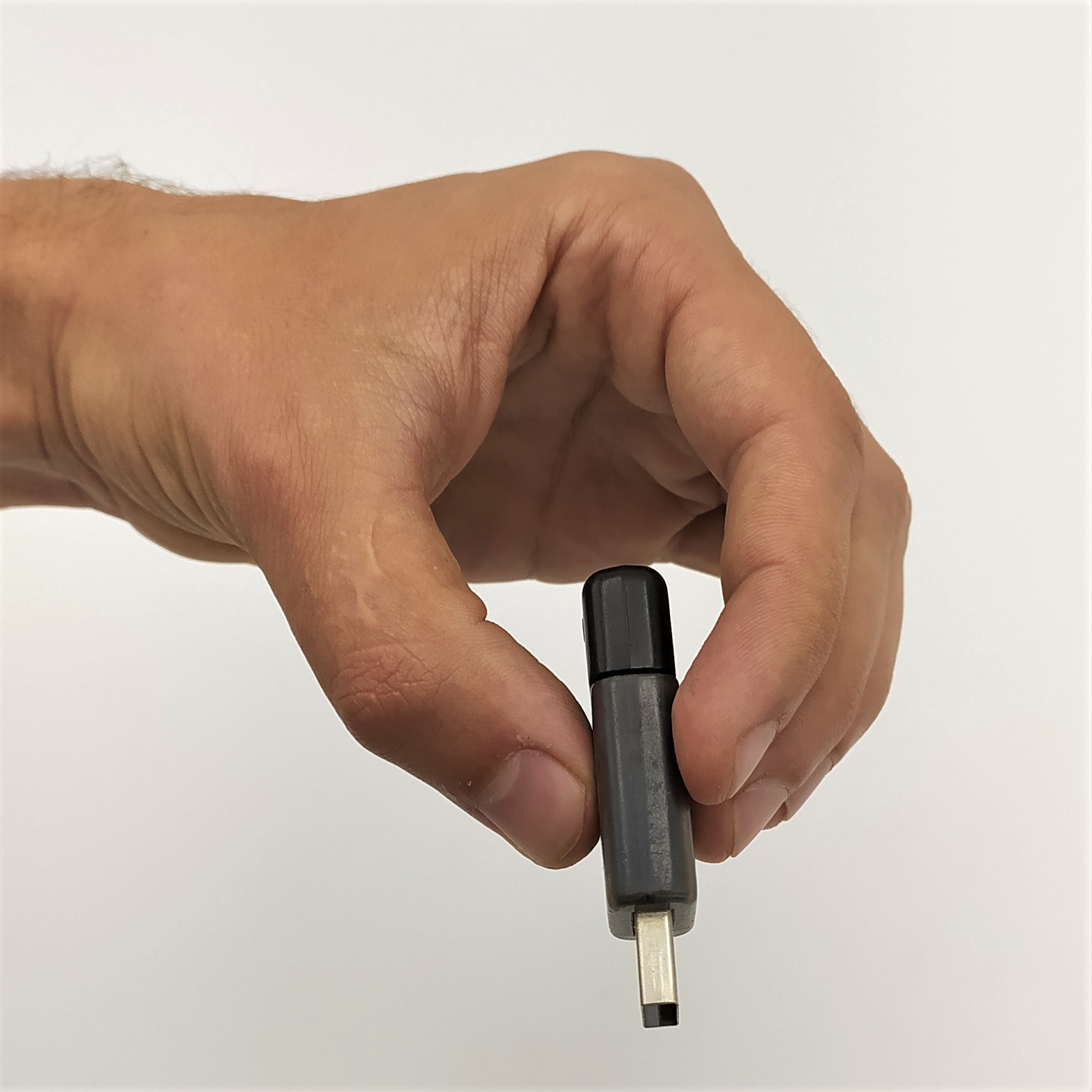}
    \end{subfigure}
    \begin{subfigure}[b]{0.35\textwidth}
        \centering
        \includegraphics[width=\textwidth]{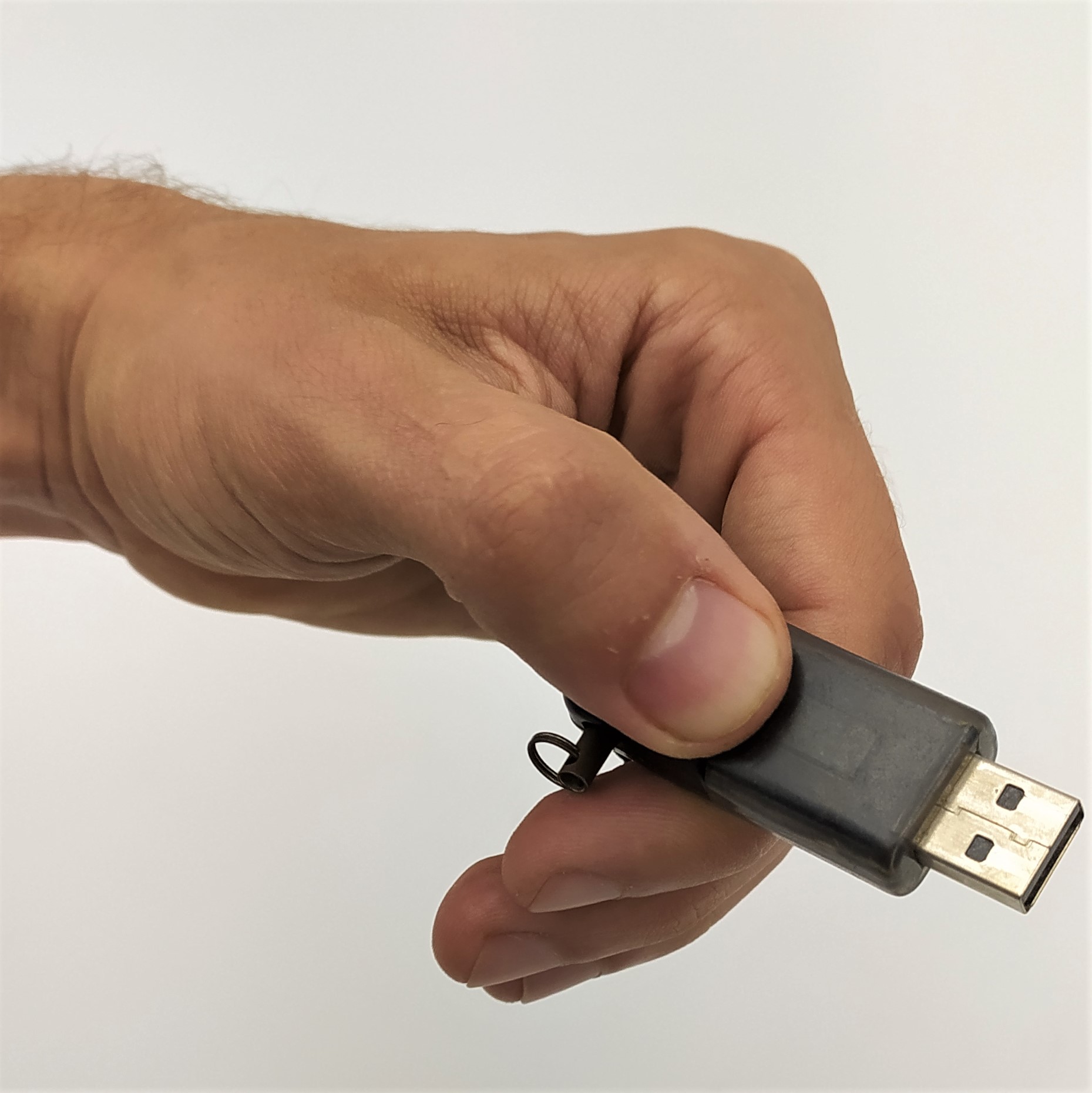}
    \end{subfigure}
    \caption{Maintien d'une clé USB en prise opposée (gauche), également appelée prise tridigitale ou prise tripode, et en prise latérale (droite).}
    \label{fig:prises}
\end{figure}

Partant du constat que plus de 90\% de la population mondiale n’a accès à aucune aide technique selon l'OMS\footnote{\url{https://www.who.int/news-room/fact-sheets/detail/assistive-technology}}, nous proposons ici une nouvelle architecture de main myoélectrique Open Source\footnote{Fichiers source disponibles sur GitHub~: \url{https://github.com/orthopus/01-myohand}} permettant son accessibilité et sa robustesse tout en conservant de bonnes performances et fonctionnalités, en collaboration avec le projet BionicoHand\footnote{Description du projet~: \url{https://bionico.org/cahier-des-charges-technical-specs/}}.
Dans cet article, une analyse cinématique des prothèses tridigitales est proposée, afin d'identifier les avantages d'un tel mécanisme. Une nouvelle cinématique de main est ensuite présentée. Celle-ci est inspirée d'une prothèse tridigitale, qui réalise seulement la prise opposée (ou prise tridigitale), à laquelle est ajoutée un mouvements d'antépulsion et de rétropulsion du pouce, permettant alors également la réalisation de la prise latérale. Un exemple de prise opposée et de prise latérale est montré sur la figure~\ref{fig:prises}. Le mouvement de rétropulsion est passif, l'utilisateur vient indexer manuellement le pouce en position latérale ou opposée. Pour la plupart des gestes du quotidien, l'indépendance des doigts n'est pas nécessaire \cite{montagnaniindependent2016}, et parmi les cinq gestes de la main les plus réalisées, trois pourraient être réalisées en prise opposée et deux autres en prise latérale \cite{bullockgrasp2013}. Le gain de fonctionnalité est donc important au regard de la complexification du système. On notera que la main Michelangelo d'Ottobock, une prothèse commercialisée utilisant deux actionneurs \cite{beltermechanical2013}, réalise uniquement ces deux prises.

Pour permettre ce nouveau mouvement, une transmission du mouvement avec deux câbles antagonistes vient remplacer la bielle rigide utilisée dans les prothèses tridigitales, entraînant l’ouverture et la fermeture du pouce en même temps que les doigts supérieurs.

En premier lieu, une analyse de la cinématique des prothèses tridigitales est proposée dans la section 2. Dans la section 3, les critères et la méthode de placement d'une nouvel axe de liaison sur le pouce sont détaillées. La section 4 se concentre sur la conception d'un système à câbles pour la transmission des efforts en détaillant les performances obtenues comparativement à la section 2. La fabrication d'un prototype et son évaluation par un utilisateur sont proposées en section 5. Finalement, la section 6 offre une conclusion et les travaux envisagés pour la suite.

\section{Analyse de la Cinématique des Prothèses de Main Tridigitales}
\medskip

Développées et utilisées à partir des années 1960 \cite{muzumdarpowered2004}, les prothèses tridigitales ne comportent qu'un seul degré d'actionnement. De nombreuses prothèses tridigitales ont été et sont encore proposées par les différents fabricants occidentaux, parmi lesquels Ottobock\footnote{\url{https://www.ottobock.com/fr-fr/accueil}} (DMC Plus, Digital Twin, SensorHand Speed, VariPlus Speed), Steeper\footnote{\url{https://www.steepergroup.com/}} (Myo Kinisi) ou Fillauer\footnote{\url{https://fillauer.com/}} (Motion Control Hand, Centri myoelectric hand). Ces prothèses sont également développées hors du marché occidental et à des prix plus bas par d'autres entreprises, comme Motorica\footnote{\url{https://global.motorica.org/}} (INDY, Russie), Tehlin\footnote{\url{https://www.tehlin.com/}} (Benatural, Taiwan) ou Wonderful Rehabilitation\footnote{\url{https://www.wonderful-po.com/}} (Electrical Hand, Chine).

Les doigts sont rigides au niveau des articulations interphalangiennes, ce qui limite l'anthropomorphisme et la stabilité des prises, mais qui augmente considérablement la simplicité et la robustesse du système. Les deux doigts supérieurs réalisant la prise tridigitale, l'index et le majeur, sont rigidement liés entre eux. De plus, l'annulaire et l'auriculaire peuvent être rigidement fixés aux deux autres doigts supérieurs, ou simplement ajoutés de manière cosmétique au niveau du gant de protection. 

\begin{figure}[!ht]
    \centering
    \begin{subfigure}[b]{0.5\textwidth}
        \centering
        \includegraphics[width=\textwidth]{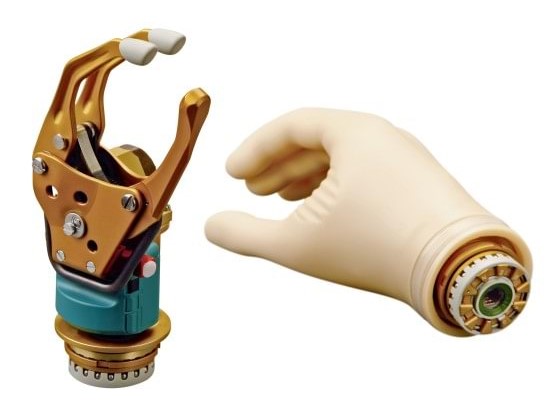}
    \end{subfigure}
    \hfill
    \begin{subfigure}[b]{0.4\textwidth}
        \centering
        \includegraphics[width=\textwidth]{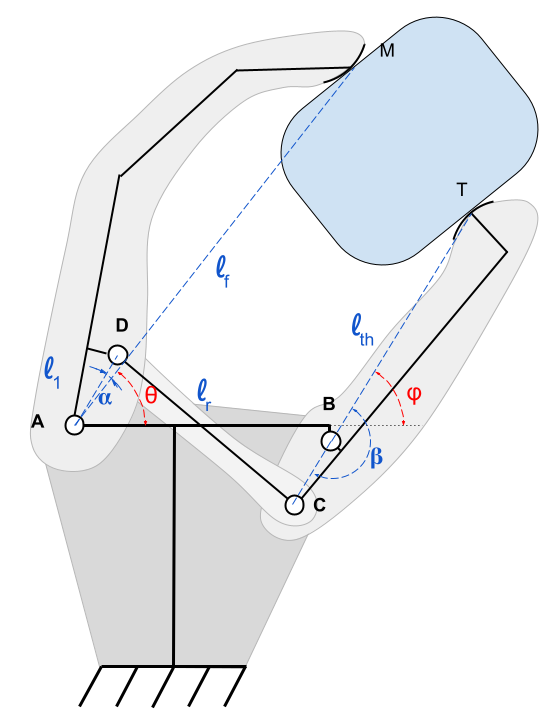}
    \end{subfigure}
    \caption{Prothèse de main tridigitale. Gauche~: modèle VariPlus Speed, Ottobock (\textit{source~: ottobock.de}). Droite~: Schéma cinématique de la prothèse.} 
    \label{fig:variplusspeed}
\end{figure}

La figure~\ref{fig:variplusspeed} présente un modèle de prothèse couramment vendu en France, ainsi qu'un schéma cinématique présentant son fonctionnement. Afin de réaliser le mouvement de flexion des doigts pour venir effectuer la prise, le pouce est entraîné par une bielle fixée sur les doigts supérieurs, eux-mêmes entraînés par la motorisation via un engrenage droit partiel.

Parmi les caractéristiques décrivant les performances d'une prothèse de main, la force de préhension et le temps de fermeture de la main sont simples à quantifier et significatives pour l'utilisateur. Ces données sont hautement liées à la cinématique du mécanisme retenu. Avec les notations de la figure~\ref{fig:variplusspeed}, la force de préhension correspond à la force appliquée par les doigts supérieurs sur l'objet au point M (égale à la force appliquée par le pouce sur l'objet au point T) pendant la phase de serrage, à vitesse nulle ou très faible. Le temps de fermeture est mesuré entre la position ouverte et la position fermée de la main pour une vitesse de rotation des doigts supérieurs $\omega_{in}$ donnée. Ce temps dépend par ailleurs de la course maximale de la main, exprimée par la distance maximale entre les doigts $d_{max}(MT)$, dont il convient de vérifier qu'elle est suffisante pour les besoins des utilisateurs. On notera que les points de contact peuvent varier légèrement en fonction de l'objet serré, mais sont considérés fixes dans cette analyse.

\begin{figure}[!ht]
\centering
\includegraphics[scale=0.75]{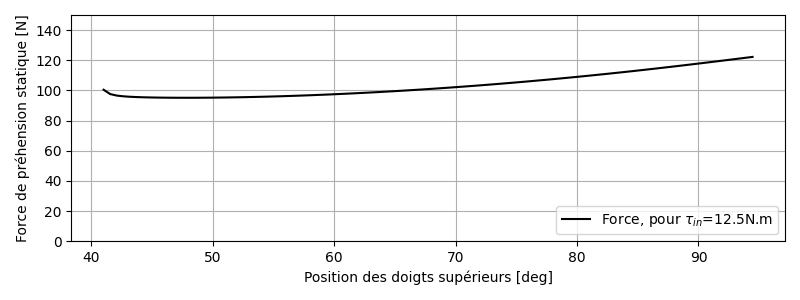}
\caption{Cinématique d'une prothèse tridigitale~: Force de préhension en fonction de la course des doigts supérieurs}
\label{fig:DMCforcespeed}
\end{figure}

Une simulation statique a été réalisée à partir des valeurs du tableau~\ref{tab:valeursgeometriques}, où $\tau_{in}$ est le couple appliqué par la motorisation sur les doigts supérieurs. Cette simulation permet d'obtenir l'évolution de la force à l'équilibre statique selon la course des doigts supérieurs, comme présenté sur la figure~\ref{fig:DMCforcespeed}. Le temps de fermeture est de 0.37 secondes pour une vitesse d'entrée $\omega_{in}~=~\ang{150}/\text{s}$, avec une ouverture maximale des doigts de $d_{max}(MT)$~=~110~mm.

\begin{table}[!ht]
    \centering
    \begin{tabular}{c|c|c|c|c|c|c|c}
        $l_{f}$ & $l_{th}$ & $l_1$ & $l_2$ & $l_r$
        & $d(AB)$ &
        $\alpha$ & $\beta$ \\
        \hline
        75~mm & 52~mm & 14~mm & 13~mm & 40~mm & 43.8~mm & $\ang{8}$ & $\ang{180}$ 
    \end{tabular}
    \caption{Valeurs mesurées des paramètres géométriques sur une prothèse VariPlus Speed (taille 7~$3/4$)}
    \label{tab:valeursgeometriques}
\end{table}

On constate que la force de la prothèse n'est pas constante sur la course du mécanisme. Le principal impact de cette non-linéarité est le surdimensionnement du moteur. En effet, pour être certain d'appliquer au minimum une force donnée sur un objet peu importe sa taille, il faut alors dimensionner le couple moteur sur le pire cas, c'est-à dire pour le minimum de la force sur l'ensemble de la course. Avec une force non constante, il peut aussi être plus difficile pour l'utilisateur de maîtriser la force de serrage si la prothèse ne dispose pas d'un asservissement en force. L'analyse montre que la force du mécanisme varie d'environ 25\% sur la course des doigts. Ce niveau de variation semble toutefois acceptable, au regard des diverses contraintes de conception, de la sensibilité de contrôle de l'utilisateur, et de la popularité et de la longévité des prothèses tridigitales. Ces résultats peuvent donc servir de référence pour comparer d'autres mécanismes.

\section{Ajout d'un Axe de Rétropulsion Passif sur le Pouce}

Dans le but d'augmenter l'anthropomorphisme et la fonctionnalité de prothèse, il est souhaitable de disposer d'une prise latérale. Il est alors nécessaire d'ajouter un degré de liberté hors plan pour permettre le mouvement de rétropulsion - antépulsion du pouce. Le passage de la prise opposée à la prise latérale est passif (non motorisé) et effectué par l'utilisateur lui-même en bougeant le pouce de la prothèse. Ce type de changement de prise est implémenté sur certaines prothèses commercialisées comme l'i-Limb Access\footnote{\url{https://www.ossur.com/en-us/prosthetics/arms/i-limb-access}} d'Össur. La cinématique de la prothèse est donc modifiée.

La première étape consiste à définir géométriquement l'emplacement de la nouvelle liaison. Cette liaison est placée entre le châssis et la liaison de flexion du pouce. L'emplacement de cette liaison pivot est complètement décrit par quatre paramètres indépendants, et deux positions angulaires de cette liaison sont choisies pour correspondre aux deux prises souhaitées. La prise opposée est conservée par rapport aux prothèse tridigitales avec le pouce dans le même plan que les doigts supérieurs. La prise latérale est placée avec un angle de rétropulsion de $\ang{90}$ par rapport aux doigts supérieurs afin que le pouce vienne serrer l'objet perpendiculairement à la surface plane de l'index.

Plusieurs critères permettent de déterminer les quatre paramètres décrivant le placement de cette liaison~:
\begin{itemize}
    \item les positions ouvertes (pour l'angle maximal de flexion) du pouce dans les prises latérales et opposées doivent permettre d'atteindre les ouvertures spécifiées (100~mm en prise opposée et 35~mm en prise latérale dans ce projet);
    \item les positions fermées (pour l'angle  minimal de flexion) du pouce dans les prises latérales et opposées doivent correspondre à la position de contact entre le pouce et les doigts supérieurs;
    \item la prothèse doit être la plus anthropomorphique possible, en essayant par exemple de rapprocher la base du pouce du plan de la paume de main en position latérale;
    \item l'emplacement de la liaison doit rester dans le volume initial d'une prothèse tridigitale, afin de rester proche de l'esthétique initiale et de pouvoir couvrir le mécanisme avec des gants de protection en silicone disponibles pour les prothèses tridigitales.
\end{itemize}

\begin{figure}[!ht]
\centering
\includegraphics[width=0.4\linewidth]{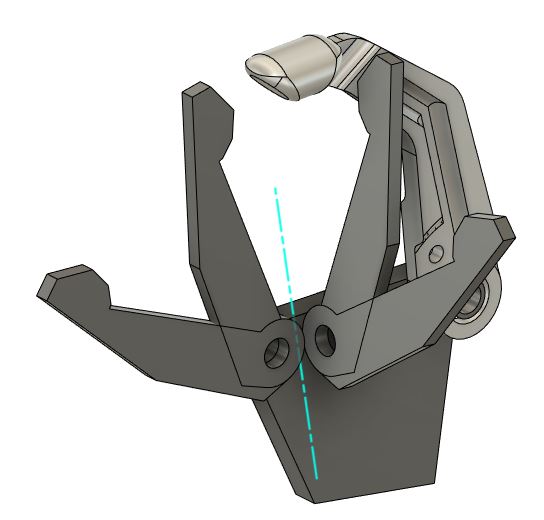}
\caption{Visualisation et validation de l'axe de rétropulsion du pouce par placement des quatre positions extrêmes~: opposée ouverte, opposée fermée, latérale ouverte, latérale fermée.}
\label{fig:positionspouce}
\end{figure}

Afin de répondre au mieux à ces critères divers, une optimisation itérative par essai-erreur a été effectuée en utilisant un logiciel de CAO. Les quatre positions extrêmes du pouce sont générées en fonction des paramètres de liaisons sélectionnés, permettant la validation de l'ensemble des critères par le concepteur, comme présenté sur la figure~\ref{fig:positionspouce}.

\section{Conception d'une Transmission par Câble}

\subsection{Motivations et Choix des Câbles}

Le mécanisme à quatre barres utilisé dans les prothèses tridigitales est simple à fabriquer, et donc potentiellement accessible et fiable. Malheureusement, il n'est compatible qu'avec des mouvements plans, contrairement au mouvement souhaité en prise latérale. Une première solution, consistant en l'utilisation de bielles à rotules, a été mise de côté dans un premier temps par la difficulté de concevoir une telle liaison à la fois compacte, peu chère et capable de supporter des efforts importants. Une deuxième solution, retenue dans cette étude, est l'utilisation de ``tendons'' (câbles ou cordes textiles). Cette solution est largement répandue dans la conception de prothèse de main \cite{bennettmultigrasp2015, kontoudisopen-source2015, laffranchihannes2020}, permettant la transmission d'efforts importants dans des volumes faibles, mais également une grande liberté d'intégration pour conserver les formes anthropomorphiques de la main.

Les câbles transmettent cependant les efforts dans une seule direction. Seul le mouvement de fermeture (flexion), soumis à des efforts plus importants lors du serrage que le mouvement d'ouverture (extension), est étudié dans un premier temps.

\subsection{Cinématique et Intégration du Câble Fléchisseur}

\begin{figure}[h!]
\centering
\includegraphics[width=1.0\linewidth]{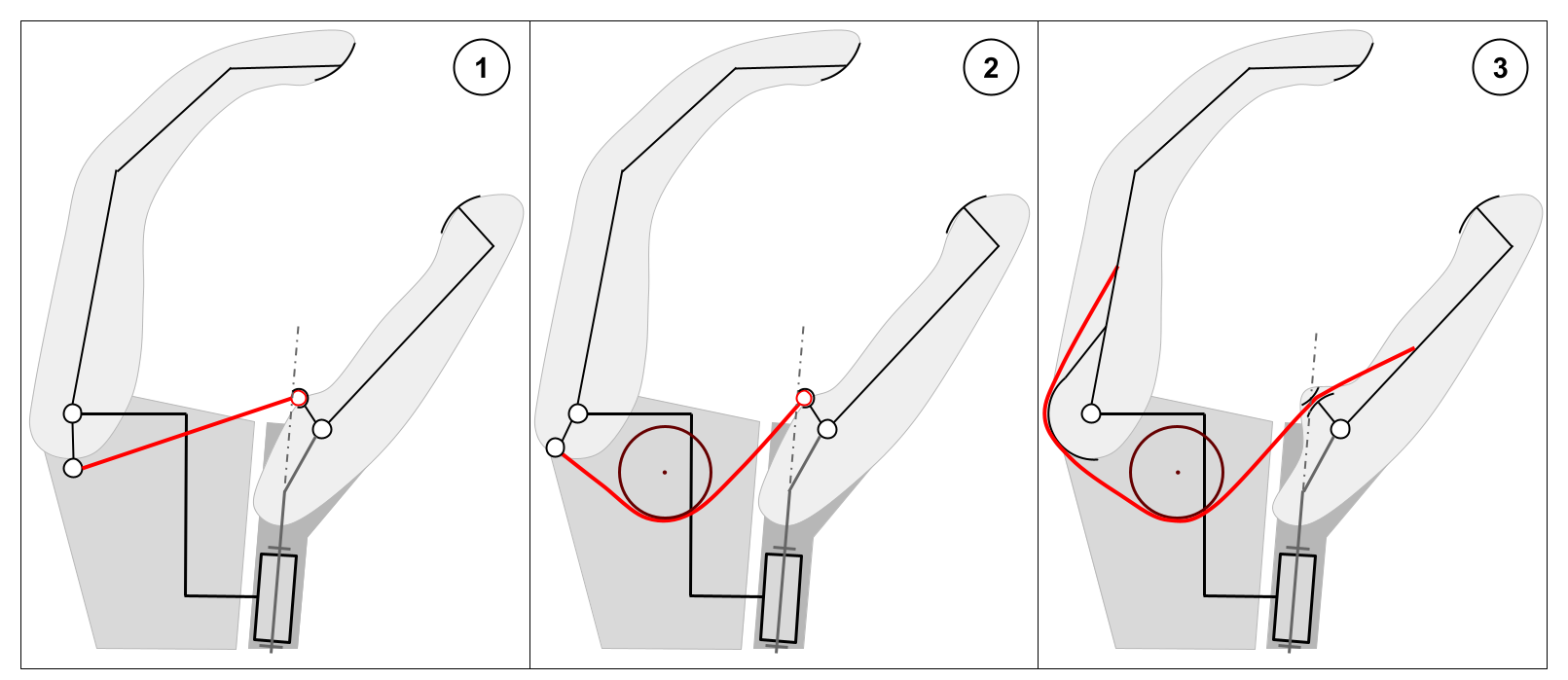}
\caption{Différentes itérations de la transmission à câble~: (1) remplacement d'une bielle rigide par un câble, (2) ajout d'une poulie de renvoi d'angle pour augmenter les bras de levier sur les doigts, (3) ajout de surface d'enroulement sur les doigts pour rendre constant les bras de levier.}
\label{fig:iterations_cable}
\end{figure}

La figure~\ref{fig:iterations_cable} présente les différentes itérations de l'intégration du câble. Lors de la première itération, la précédente solution de mécanisme à quatre barres sert de point de départ, et la bielle est remplacée par un câble. Le sens de la bielle est inversée (fixée en dessous du point de pivot des doigts supérieurs et au dessus du point de pivot du pouce) pour fonctionner en traction pendant la fermeture. 
Le câble est relié via des oeillets aux doigts supérieurs et au pouce, modélisés dans l'analyse par des liaisons rotules.
La position de la liaison rotule entre le pouce et le câble est très proche de l'axe de rétropulsion du pouce, lors de la fermeture du mécanisme. De cette manière, le câble conserve la même position après un mouvement de rétropulsion du pouce. Ainsi, le câble peut actionner le pouce en prise latérale comme en prise opposée.

Dans une deuxième itération, on utilise une poulie pour orienter la force appliquée au niveau du pouce davantage parallèlement à l'axe de rétropulsion du pouce. En effet, si la force exercée par le câble est perpendiculaire à l'axe de rétropulsion du pouce, le câble pourra faire tourner le pouce en prise opposée. En prise latérale, le câble deviendra parallèle à l'axe de flexion du pouce et le mouvement ne pourra être transmis. L'ajout de cette poulie permet donc de transmettre au mieux le mouvement en prise latérale comme en prise opposée.

Après de premières simulations, une troisième itération a permis de supprimer la singularité du mécanisme~: lors de l'ouverture complète des doigts supérieurs, ces doigts s'alignent avec le câble, diminuant considérablement le bras de levier. Autour de cette position, un faible couple appliqué sur les doigts supérieurs peut appliquer une tension très importante dans le câble, et dépasser les limites mécaniques du câble ou des liaisons. Ce comportement est également synonyme de non-linéarité, comme dans la transmission originale à bielle rigide. L'ajout de surfaces d'enroulement est alors une solution efficace. Le rapport entre la tension du câble et le couple exercé sur chaque doigt devient beaucoup plus constant.

\subsection{Validations Cinématique et Statique}

\begin{figure}[!ht]
\centering
\includegraphics[scale=0.75]{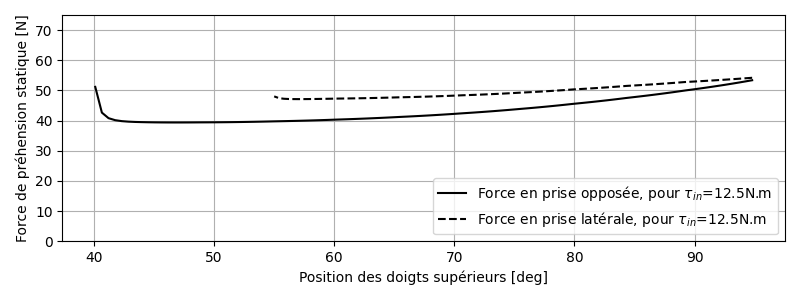}
\caption{Nouvelle cinématique~: Force de préhension en fonction de la course des doigts supérieurs, en prise latérale et opposée.}
\label{fig:Myohandforcespeed}
\end{figure}

De la même façon que l'analyse de la prothèse tridigitale, des simulations cinématiques et statiques ont été  effectuées. L'évolution de la force de préhension au cours de la course est présentée sur la figure~\ref{fig:Myohandforcespeed}. La variation de la force est de l'ordre de 25\%, soit un niveau de variation équivalent à celui des prothèses tridigitales. On observe que la position extrême fermée des doigts supérieurs n'est pas égale en prise opposée et en prise latérale, contrairement à l'hypothèse formulée lors du placement de liaison de rétropulsion du pouce. Le point d'accroche évolue autour de l'axe de rétropulsion, mais n'est pas confondu avec. L'angle de flexion du pouce qui n'est pas constant lors du changement de prise. 

Cela est sans conséquence sur le fonctionnement du mécanisme, tant que le pouce vient bien en contact du côté de l'index, ce qui a été validé. Le temps de fermeture est conservé à moins de 0,4 secondes pour une vitesse d'entrée $\omega_{in} = \ang{150}/\text{s}$, l'ouverture maximale des doigts est de 100~mm en prise opposée et 40~mm en prise latérale.

\subsection{Intégration d'un Câble Extenseur}

Comme évoqué précédemment, un câble ne peut transmettre d'effort que dans un sens. Plusieurs solutions peuvent alors être envisagées pour assurer l'ouverture du pouce en même temps que l'ouverture des doigts supérieurs.

La première solution est l'utilisation d'un ressort de rappel. Ce ressort se tend lors de la fermeture, et rappel alors le pouce vers l'extérieur lorsque les doigts supérieurs s'ouvrent et que le câble se détend. Ce ressort doit être suffisamment rigide pour déformer le gant de protection jusqu'à l'ouverture complète. Une telle solution est utilisée dans la prothèse Michelangelo \cite{puchhammer2007-michel}. Elle est simple à intégrer mais augmente le couple nécessaire à la fermeture de la prothèse pour charger le ressort. Un moteur plus puissant doit alors être sélectionné, amenant des problèmes de coût, de consommation énergétique et de masse.

L'ajout d'un câble antagoniste est une deuxième solution. Ce câble extenseur lie alors le pouce et les doigts supérieurs de sorte à être tendu lors du mouvement d'ouverture. Cette solution présente l'avantage de ne pas ajouter de couple lors de la fermeture. Il est cependant important de remarquer que le système devient hyperstatique. Pour chaque position des doigts supérieurs, il devient alors primordial que la position du pouce obtenue en tendant le câble extenseur soit égale à celle obtenue en tendant le câble fléchisseur. Cette contrainte peut être légèrement relâchée en considérant le comportement élastique des câbles.

Cette solution a été retenue et implémentée dans le développement du prototype. L'intégration du câble extenseur est similaire à celle du câble fléchisseur. En partant du mécanisme à quatre barres, la bielle rigide est remplacée par un câble. Ce câble travaille bien en traction lors de l'ouverture. Des surfaces d'enroulement sont ajoutées sur le pouce pour rendre constant le bras de levier entre le câble extenseur et le pouce. La bonne intégration de ce câble, dépendante de l'élasticité du câble et des jeux de montage, a été validée expérimentalement.

\section{Expérimentation et Évaluation}

\subsection{Présentation du Prototype}

Un prototype a été conçu pour valider cette nouvelle cinématique. Le châssis et le corps des doigts ont été fabriqués en aluminium par fabrication additive, permettant des itérations rapides de prototypage. Un moteur à courant continu Portescap 22N78 avec son réducteur K24 est utilisé, contrôlé par une Arduino Nano Every et un Driver Pololu VNH5019. Le reste de la transmission est constitué d'un engrenage et vis sans fin irréversible, et de deux étages d'engrenages droits. Cette transmission permet d'atteindre une vitesse de rotation des doigts supérieurs de $\ang{150}/\text{s}$ (fermeture en moins de 0,4 secondes) et un couple de 6~N.m.

\begin{figure}[!ht]
\centering
\includegraphics[width=0.8\linewidth]{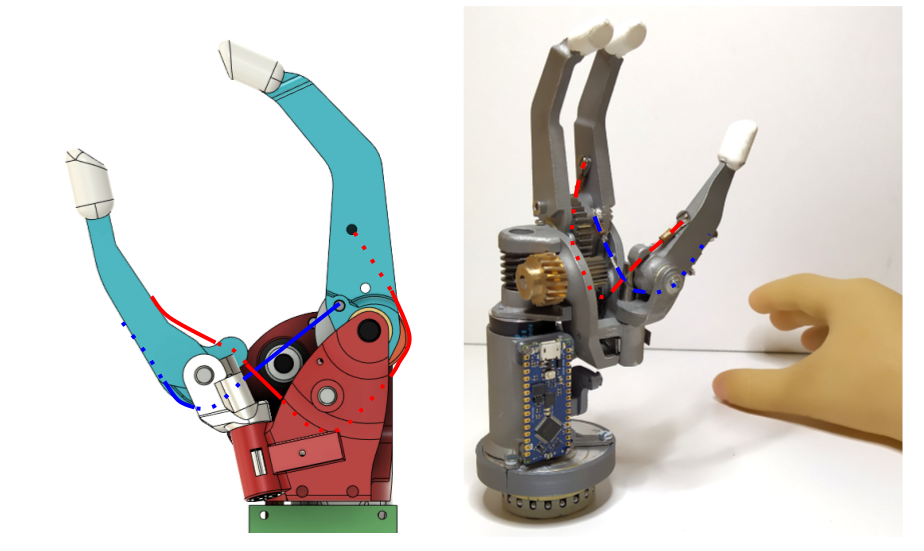}
\caption{Chemin des câbles dans la prothèse avec en rouge le câble fléchisseur (fermeture) et en bleu le câble extenseur (ouverture). Gauche~: vue CAO en prise opposée. Droite~: photo du prototype en prise latérale.}
\label{fig:proto}
\end{figure}

Deux types de câbles ont été testés. Les cordes textiles (type Dyneema) sont adaptées à de grands efforts de traction et des rayons de courbures faibles, mais leur importante élasticité ne permet pas de respecter la cinématique initialement conçue. Des câbles en acier de diamètre 1~mm sont finalement retenus. Ces câbles sont beaucoup plus rigides, mais un rayon minimum de courbure plus important est nécessaire pour assurer une bonne durée de vie. Des rondelles de calages permettent de régler les jeux au montage, au niveau des oeillets. Le résultat de l'intégration est présenté sur la figure~\ref{fig:proto}.

L'indexation de l'axe de rétropulsion du pouce est réalisée par la présence de deux rainures dans l'axe. Une lame de ressort permet alors de bloquer l'axe dans sa position via un ergot. Au-delà d'un certain couple appliqué par l'utilisateur, l'ergot sort de la rainure, et le pouce peut être placé en position opposée ou latérale.

Le prototype est utilisé avec un gant en silicone pour prothèse tridigitale. Le pouce est moulé dans sa position en prise opposée, mais est suffisamment élastique pour réaliser la prise latérale.

\subsection{Évaluation du Prototype}

Dans une démarche de co-conception, un utilisateur expérimenté, équipé au quotidien d'une prothèse Michelangelo et d'une VariPlus Speed (Ottobock), a été interrogé sur le prototype réalisé.

Le temps de fermeture de 0,4 secondes convient à l'utilisateur. Le changement de position du pouce peut être réalisé soit par la main opposée ou soit en utilisant l'environnement comme obstacle. Contrairement à la prothèse quotidienne de l'utilisateur, la rétropulsion n'est toutefois pas motorisée, ce qui demande une adaptation et une acceptation.

La première remarque notée par l'utilisateur est l'esthétique trop peu anthropomorphique de la prothèse. En position latérale, la main ne semble pas ``naturelle'' et peut retenir l'attention de l'entourage.

Il a également été noté une mauvaise stabilité de la prise latérale~: le pouce se repositionne parfois de lui-même en position opposée lorsqu'un objet est serré. Cela semble principalement dû à l'élasticité du gant en silicone, qui exerce un couple au niveau de la liaison de rétropulsion du pouce, proche du couple d'indexation nécessaire au changement de position du pouce. Augmenter davantage le couple de rétropulsion pourrait rendre difficile le déplacement du pouce de la position opposée à la position latérale.

Enfin, la force de préhension disponible, mesurée à environ 40N, est également trop faible pour maintenir les objets avec aisance et fiabilité (68N sont nécessaires pour réaliser la plus grande partie des tâches quotidiennes \cite{Weirdesign2004}). 

\section{Conclusion et Perspectives}

Cet article propose une nouvelle cinématique de prothèse de main, et analyse les performances associées à cette cinématique au regard des performances des prothèses tridigitales. Les différentes étapes de la conception de la transmission par câble y sont détaillées. Ce travail a pour objectif de proposer une prothèse simple à fabriquer, à utiliser, à réparer, mais également performante et adaptée aux besoins de tous les jours des utilisateurs.

Un prototype a été conçu et fabriqué, permettant de confronter les idées proposées aux difficultés techniques de fabrication, montage, jeux et élasticité des différentes pièces. Cela a surtout permis la confrontation des idées à un utilisateur expérimenté, permettant de tester l'acceptabilité de l'esthétique et des prises de la prothèse.

La prothèse proposée valide l'intérêt de l'ajout d'une liaison de rétropulsion passive, plus simple qu'une rétropulsion motorisée et plus fonctionnelle qu'une prothèse tridigitale. Il valide aussi la possibilité d'utiliser des câbles en acier pour la réaliser d'une telle transmission avec des forces de serrage modérées.

Un point faible de ce prototype est son esthétique. Il serait intéressant dans de futurs travaux de s'éloigner de la cinématique des prothèses tridigitales pour proposer des mouvements plus anthropomorphiques. D'autres études sont en cours pour augmenter le couple disponible au niveau des doigts en limitant la consommation énergétique de la prothèse, notamment en utilisant un réducteur à deux vitesses \cite{butindesign2022} et une transmission irréversible. L'utilisation de câbles pour transmettre des efforts plus importants pourra alors être approfondie. L'étude du rendement d'une telle transmission à câble pourra également être quantifiée.


\bibliographystyle{acm}
 \bibliography{myohand_kinematics.bib}

\begin{thebibliography}{10}

\bibitem{beltermechanical2013}
{\sc Belter, J.~T., Segil, J.~L., Dollar, A.~M., and Weir, R.~F.}
\newblock Mechanical design and performance specifications of anthropomorphic
  prosthetic hands: {A} review.
\newblock {\em The Journal of Rehabilitation Research and Development 50}, 5
  (2013), 599--618.

\bibitem{bennettmultigrasp2015}
{\sc Bennett, D.~A., Dalley, S.~A., Truex, D., and Goldfarb, M.}
\newblock A {Multigrasp} {Hand} {Prosthesis} for {Providing} {Precision} and
  {Conformal} {Grasps}.
\newblock {\em IEEE/ASME Transactions on Mechatronics 20}, 4 (Aug. 2015),
  1697--1704.

\bibitem{bullockgrasp2013}
{\sc Bullock, I.~M., Zheng, J.~Z., De~La~Rosa, S., Guertler, C., and Dollar,
  A.~M.}
\newblock Grasp {Frequency} and {Usage} in {Daily} {Household} and {Machine}
  {Shop} {Tasks}.
\newblock {\em IEEE Transactions on Haptics 6}, 3 (July 2013), 296--308.

\bibitem{butindesign2022}
{\sc Butin, C., Chablat, D., Aoustin, Y., and Gouaillier, D.}
\newblock {Design of a Two-Speed Load Adaptive Variable Transmission for
  Energetic Optimization of an Accessible Prosthetic Hand}.
\newblock {\em Journal of Mechanisms and Robotics 15}, 1 (04 2022).

\bibitem{kontoudisopen-source2015}
{\sc Kontoudis, G.~P., Liarokapis, M.~V., Zisimatos, A.~G., Mavrogiannis,
  C.~I., and Kyriakopoulos, K.~J.}
\newblock Open-source, anthropomorphic, underactuated robot hands with a
  selectively lockable differential mechanism: {Towards} affordable prostheses.
\newblock In {\em 2015 {IEEE}/{RSJ} {International} {Conference} on
  {Intelligent} {Robots} and {Systems} ({IROS})\/} (Hamburg, Germany, Sept.
  2015), IEEE, pp.~5857--5862.

\bibitem{laffranchihannes2020}
{\sc Laffranchi, M., Boccardo, N., Traverso, S., Lombardi, L., Canepa, M.,
  Lince, A., Semprini, M., Saglia, J.~A., Naceri, A., Sacchetti, R., Gruppioni,
  E., and De~Michieli, L.}
\newblock The {Hannes} hand prosthesis replicates the key biological properties
  of the human hand.
\newblock {\em Science Robotics 5}, 46 (Sept. 2020), eabb0467.

\bibitem{montagnaniindependent2016}
{\sc Montagnani, F., Controzzi, M., and Cipriani, C.}
\newblock Independent {Long} {Fingers} are not {Essential} for a {Grasping}
  {Hand}.
\newblock {\em Scientific Reports 6}, 1 (Dec. 2016), 35545.

\bibitem{muzumdarpowered2004}
{\sc Muzumdar, A.}, Ed.
\newblock {\em Powered {Upper} {Limb} {Prostheses}: {Control}, {Implementation}
  and {Clinical} {Application}}.
\newblock Springer Berlin Heidelberg, Berlin, Heidelberg, 2004.

\bibitem{puchhammer2007-michel}
{\sc Puchhammer, G.}
\newblock Hand prosthesis with fingers that can be aligned in an articulated
  manner, Oct. 2007.
\newblock WO2007076765A3.

\bibitem{Weirdesign2004}
{\sc Weir, R.~F.}
\newblock Design of {Artificial} {Arms} and {Hands} for {Prosthetic}
  {Applications}.
\newblock In {\em Standard {Handbook} of {Biomedical} {Engineering} \&
  {Design}}, digital engineering library mcgraw-hill~ed. 2004, pp.~32.1 --
  32.61.

\end{thebibliography}
\end{document}